# Human Activity Recognition models using Limited Consumer Device Sensors and Machine Learning

Wei Zhong Tee
Student, Department of Computer Science
University of Wisconsin Eau-Claire
Eau Claire, United States
teewz1076@uwec.edu

Rushit Dave
Assistant Professor, Department of Computer Science
University of Wisconsin Eau-Claire
Eau Claire, United States
daver@uwec.edu

Jim Seliya
Assistant Professor, Department of Computer Science
University of Wisconsin Eau-Claire
Eau Claire, United States
seliyana@uwec.edu

Mounika Vanamala
Assistant Professor, Department of Computer Science
University of Wisconsin Eau-Claire
Eau Claire, United States
vanamalm@uwec.edu

*Abstract*—Human activity recognition has grown in popularity with its increase of applications within daily lifestyles and medical environments. The goal of having efficient and reliable human activity recognition brings benefits such as accessible use and better allocation of resources; especially in the medical industry. Activity recognition and classification can be obtained using many sophisticated data recording setups, but there is also a need in observing how performance varies among models that are strictly limited to using sensor data from easily accessible devices: smartphones and smartwatches. This paper presents the findings of different models that are limited to train using such sensors. The models are trained using either the k-Nearest Neighbor, Support Vector Machine, or Random Forest classifier algorithms. Performance and evaluations are done by comparing various model performances using different combinations of mobile sensors and how they affect recognitive performances of models. Results show promise for models trained strictly using limited sensor data collected from only smartphones and smartwatches coupled with traditional machine learning concepts and algorithms.

## I. Introduction

Machine learning applications have expanded rapidly over the past few decades [1]. Uses for machine learning vary from many topics, including continuous [2] and user [3] authentication schemes [4], cybersecurity [5] and IoT applications [6]. Another area of application of machine learning techniques is human activity recognition. Human activity recognition using machine learning is significant because it aids in health and lifestyle monitoring [7] and for clinical usage such as disease tracking and health indicator [8] tracking. In the past, many of these models required tedious models with specialized monitoring and data collection equipment, but more recently this can all be achieved while being accessible using inexpensive sensors and applicable using machine learning concepts.

Such factors can aid with general health monitoring of people and with right allocation of medical resources when it comes to detecting health issues in advances and determining the diagnosis and prognosis of a condition [9]. In a time where there is significant value in allocating medical resources efficiently, using machine learning concepts for human activity recognition can save lots of resources to be allocated elsewhere. If this can be coupled with the idea that human activity recognition can be reliable strictly using data collected from easily accessible devices, such as smartphones and smartwatches, human activity recognition capabilities will be able to expand beyond the expertise of those in the medical and health industries.

The goal of this research is to investigate how easily accessible sensors within mobile phones or smart watches can contribute towards machine learning based models for human activity recognition. This paper will investigate how using different sensors can affect results of human activity recognition models that are based on traditional machine learning algorithms.

## II. Related Works

Previously, a review on machine learning algorithms regarding human activity recognition was conducted. The review concluded with many models [10] performing well using support vector machine (SVM), k-nearest neighbor (kNN) and random forest (RF) algorithms for classifying or predicting human activities [11]. Studies were used in the contexts of both normal activity tracking and recognition and for medical use using patient [12] data and specialized sensors for obtaining data. Hence, the three same algorithms will also be used in this study for determining performance using different sensors

One study [13] utilized many phones' accelerometer, gyroscope, and magnetometer for identifying walking,

running, standing, sitting, walking upstairs, and walking downstairs. Phones were placed on five different smartphone positions on the body

Another study [14] utilized a custom device which consisted of low power sensors, on board memory and Bluetooth for tracking and storing data on a 9-axis orientation. The framework utilized shock aware segmentation, feature extraction and classification of activities using deep learning and machine learning concepts. Both studies utilized many devices and had a tedious [15] setup before being able to feed data into models for training and prediction, being subsequently inaccessible to many people even if they provided promising results.

### III. DATASET

The dataset used for these experiments is a publicly available dataset known as the Extrasensory dataset [16]. This dataset was chosen primarily for the fact that data was collected on everyday devices that many people will typically own. Sensors that recorded data were all on smartphones or smartwatches. Data was collected in the wild and not in any controlled environment and they are rich with contextual labels of activities. Beside more traditional labels such as "walking" or "sitting", they are data points for activity labels such as "at school" or "in a meeting". There are 60 participants in this dataset, and they vary in age, height, weight, and sex. Users participated anywhere between 3 to 28 days and provided over 300,000 samples (in minutes) of contextual data. This dataset was chosen for its candid nature and uncontrolled data collection techniques; best simulating data from sensors from non-specialized devices and environments.

Data collected from smartphone sensors include the accelerometer, gyroscope, location, audio, and audio properties (max absolute value of recorded audio before it was normalized). The accelerometer and gyroscope collected data at a rate of 40Hz and the location and audio data were recorded on environmental change detection. Data collected from smartwatch sensors include the accelerometer and compass. The watch accelerometer recorded data at a rate of 25Hz and the compass recorded data on any directional change by 1 degree.

### IV. METHODOLOGIES AND MODEL

With the initial step of data collection complete by using the publicly available extrasensory dataset, the subsequent step was to design the models that were to be used in the experiments.

Figure 1 shows the flow of data and model's design that will be used in the experiments. Data collection that was obtained through smartphones and smartwatches were marked with clean labels that were associated with activities the users were performing. As there were 60 users' data collected, 50 at random were chosen for training and the remaining 10 would be used for testing the model's performance.

The data then had to go through preprocessing before it could be fed into the model for training. The data that was preprocessed was selected based on which sensors were being to train the model. Likewise, appropriate labels had to be selected for the model's classifying ability to recognize and for the purpose of the experiment's purpose of observing performance responses based on sensor combinations. It was intentional that rather common activities with a variety of users and examples were chosen as the labels that were used in the experiment.

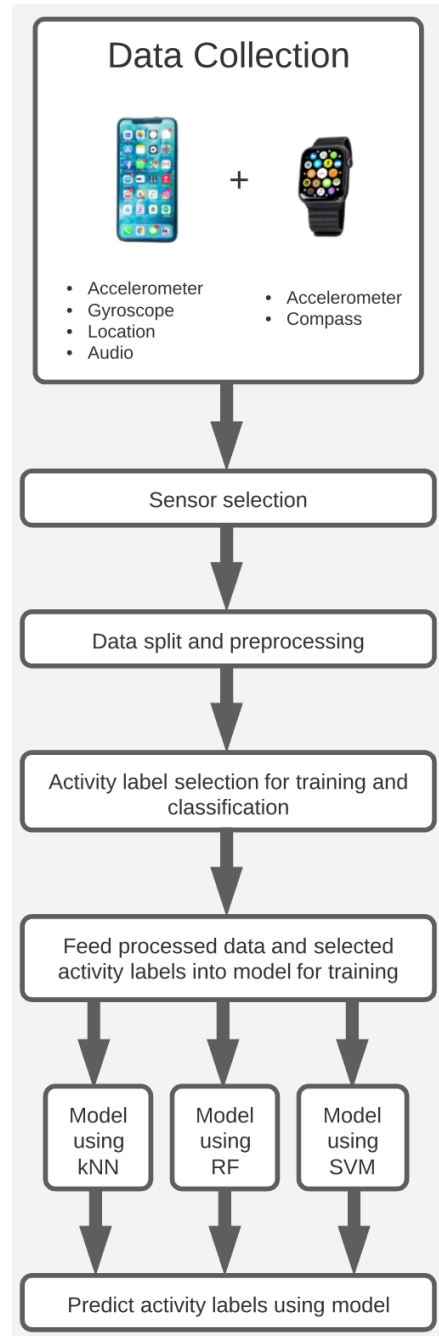

Figure 1. Flow of data throughout the experiments and model design

The models were trained using three separate algorithms: kNN, RF and SVM. These three algorithms were selected from previous works that showed good performance with human activity recognition models (See section II: Related Works). Once the models were trained, the data from the testing user group was used for determining the model's performance in classifying activity labels. The experimental results of the different sensor selections and activity labels within the separate models are observed to draw the key findings in the results from different sensor selections.

## V. Experimental Setup

A total of four experimental groups were setup for this research project. The four were divided based on a collection of different sensors and different metrics were observed to see how the use of different sensors would affect the recognitive performance of the models. The following are the groups of sensors used in the four different experiments:

I. Group 1: watch accelerometer and watch compass
II. Group 2: phone accelerometer, phone gyroscope, and phone location
III. Group 3: phone accelerometer, phone gyroscope, location, watch accelerometer and watch compass
IV. Group 4: phone accelerometer, phone gyroscope, location, watch accelerometer, watch compass, audio, and audio properties

Data from the clean labels were used from the dataset. The sensor features were extracted from each user by their associated UUID and was first normalized in the preprocessing stage. Data was centralized by subtracting the mean vector of the sensor and then normalized by dividing the centralized value by the standard deviation of the sensor's data in the event the standard deviation is not zero.

Out of the 60 user's data, 50 were randomly selected for training the model and the other 10 used for testing the predictive ability of the trained model. There was a total of five selected prediction labels for the models to predict. The five labels were walking, running, lying down, sitting, and sleeping. Of the five labels, there are some labels with more examples from recorded data and some with less. They are also common activities with little ambiguity (for example, some labels are just called "indoors"), nor are they very specific (for example, some labels are called "with friends"). For these specified reasons, the five labels were chosen to be the predictive targets for the models. Sitting, lying down, and sleeping had many examples from users, while walking had significantly fewer. Running had the least examples but was also selected to see how the model may behave around a label that had little training data. Suggestively, the classifying ability for running and walking are likely to be worse than the results of sitting, lying down and sleeping. Table I shows the distribution of the number of users and examples the selected labels have.

TABLE I. DISTRIBUTION OF LABEL USERS AND EXAMPLES

| Activity Label | Number of Users | Number of Examples |
|---|---|---|
| Sitting | 60 | 136356 |
| Lying Down | 58 | 104210 |
| Sleeping | 53 | 83055 |
| Walking | 60 | 22136 |
| Running | 26 | 1090 |

Each group's data from the listed sensor pool was normalized and fed into a model that was trained based on an SVM, kNN or RF algorithm. In short, for every group of sensors, three models were trained with the three specified machine learning algorithms. The machine learning algorithms were all imported using the python sklearn library and the model implementations were coded in python. A total of 46 features were extracted from the sensors in group 1, 52 features from group 2, 98 features from group 3, and lastly 126 features from group 4. The models were then used to predict the labels from the data split for testing and validation and were evaluated using the following metrics: accuracy, F1 score, recall, precision, and the area under the receiver operating characteristic (ROC) curve (AUC).

## VI. Experimental Results

It is worth noting a few things about the results from the experiments. Firstly, judging a model's predictive ability (especially human activity predictions) using the accuracy metric can be very misleading. A resulting high accuracy can be biased because of the higher number of non-activities examples than predicted activity examples. For example, a result can have a high accuracy, but that is because it can predict true negative results well, but often gets false positive results. It is then just as important to observe the recall (true positive rate) and the AUC (also referred to as balanced accuracy) results.

All the models performed extremely poorly for results concerning the predictive ability to classify running. This is due to the low number of users and examples provided to train the model. This is one downside of using such a context rich dataset: that the number of specific labels are so many that a broad category of activities, such as running, results in a low number of examples. The models were able to get many true negative results, but very few true positive results, resulting in all four sensor groups obtaining AUC scores of less than 0.53, while having accuracy ranges of 0.92-0.93. This is an example of the accuracy metric being misleading as it does not fully reflect the ability to predict true positive running results. The low AUC score reflects the low recall rate, of which all models had less than 0.05.

The more valuable results can be observed through the results obtained from the models and groups that were trained to predict walking, lying down, sitting, and sleeping. Group 2 often outperformed group 1 in terms of experimental metric results, but findings show that concatenation of the two groups to form group 3 does not improve results drastically. Table II shows the comparisons of results for walking from the kNN algorithm for models trained with data from group 1, group 2 and group 3.

TABLE II. kNN RESULTS OF WALKING

| Group | Accuracy | Recall | AUC | Precision | F1 Score |
|---|---|---|---|---|---|
| 1 | 0.94 | 0.16 | 0.58 | 0.64 | 0.26 |
| 2 | 0.95 | 0.41 | 0.70 | 0.69 | 0.51 |
| 3 | 0.95 | 0.39 | 0.69 | 0.71 | 0.50 |

Relationships between results for walking between group 1, group 2 and 3 were similar for all three algorithms; where there was a steady increase in all metrics to show better performance for group 2 over group 1, but not much improvement in group 3 from group 2. The lower recall value can also be attributed to the lower amount of examples walking had. On the other hand, the recall values and AUC scores for sitting, lying down and sleeping were much higher for the all the models and that can be attributed to having far more positive examples than that of walking. In fact, sitting has the best score of all the activity labels chosen and it also has the greatest number of users and examples that the model could have trained from. However, with many positive examples comes the issue of the model giving many false negative results due to the lower number of negative examples for the model to be trained with. Table III shows the results for sitting using group 4's sensors, with noticeably lower accuracy but higher recall, AUC and F1.

TABLE III. GROUP 4 SITTING RESULTS

| Algorithm | Accuracy | Recall | AUC | Precision | F1 Score |
|---|---|---|---|---|---|
| kNN | 0.70 | 0.76 | 0.70 | 0.66 | 0.67 |
| RF | 0.75 | 0.79 | 0.75 | 0.71 | 0.75 |
| SVM | 0.73 | 0.81 | 0.73 | 0.73 | 0.74 |

As previously explained, when it comes to human activity recognition, obtaining better recall score signifies better sensitivity which means the model is capable of producing true positive predictions. The results from the RF based algorithm for sitting was actually the best result from the experiments with all 4 groups, where it obtained an AUC score of 0.75. Figure 2 shows the confusion matrix for the predictive results of sitting using group 4's sensors and the RF algorithm. Performance was also very similar using the kNN and SVM algorithms. Both sets of algorithms also saw performance metrics improvements with the addition of sound data used to train the models.

Figures 3 and 4 also show the resulting confusion matrix for sitting using group 4's sensors and the kNN and SVM algorithms respectively.

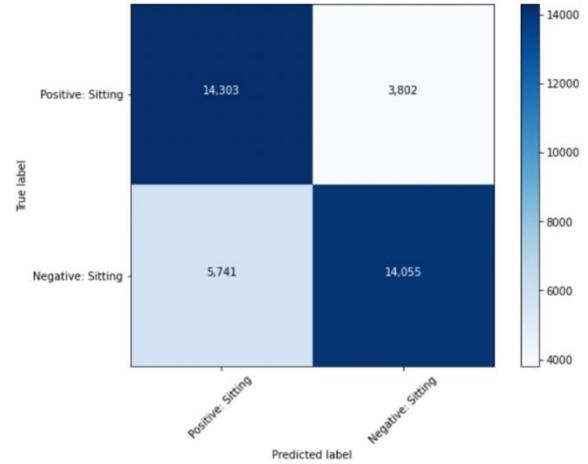

Figure 2. Confusion matrix for sitting using group 4's sensors and the RF algorithm

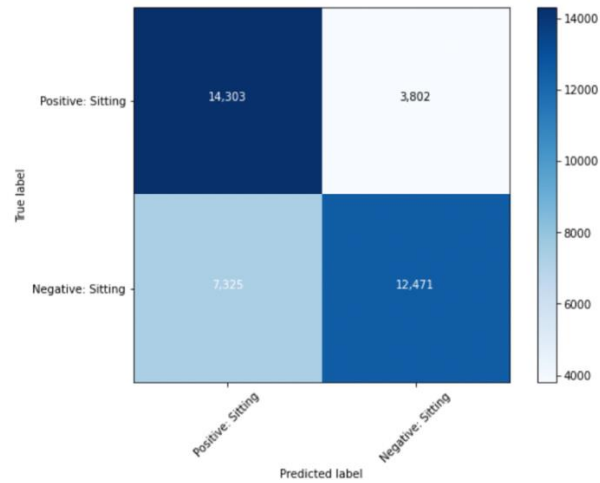

Figure 3. Confusion matrix for sitting using group 4's sensors and the kNN algorithm

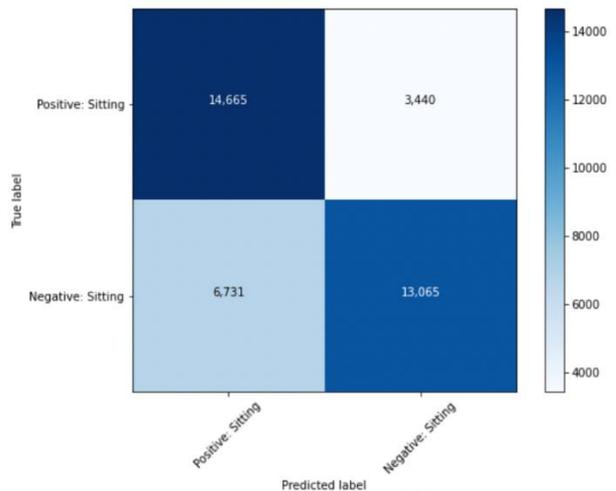

Figure 4. Confusion matrix for sitting using group 4's sensors and the SVM algorithm

The final point to note from the set of results that is interesting is the addition of sound data to group 4. The addition of sound data improved results of lying down and sleeping by a healthy margin. It makes sense that this will be the case as most users have sleep and lying down data recorded when their environments are rather quiet, creating good bias towards the labels when there is quiet sound data reported. Table IV shows how the results for sleeping improved with SVM algorithm as the model was trained with the sound data form group 4.

TABLE IV. SLEEPING RESULTS USING SVM

| Group | Accuracy | Recall | AUC | Precision | F1 Score |
|---|---|---|---|---|---|
| 3 | 0.78 | 0.38 | 0.67 | 0.75 | 0.51 |
| 4 | 0.83 | 0.48 | 0.73 | 0.86 | 0.62 |

However, using the sound data in the training of the models impacted other activity labels where sound is not as valuable towards the predictions. For example, all the metrics for the walking dropped by values of 0.1-0.3 when sound was added in group 4's experiments for all three algorithms. This suggests that the predictive ability of more passive activities, such as sleeping or lying down responds better when there is sound data to represent quieter surroundings, but more active activities such as walking suffers with worse predictive performance when sound data is introduced due to ambiguity of the bias data.

VII. DISCUSSION AND LIMITATIONS

A few conclusions can be drawn with the following experiments. Firstly, having multiple sensors that provide the same data may not always provide better results when training recognitive or predictive models. The experiments using the same algorithms but different sensors in group 1, group 2 and group 3 shows using a variety of sensors provides in group 3 not necessarily performing with better results than using data group 2. That being said, there is a healthy margin of improvement in results from group 1 (smartwatch sensors) to group 2 (smartphone sensors) as seen with the walking results and AUC score improvements. Secondly, different sensors can be either ideal or damaging towards the predictive qualities in a model depending on an activity. This was displayed through the results of using sound data to improve the sleeping and lying down classification results of the models, but in turn reducing the overall performance of the model's ability to recognize walking.

The model was also able to perform far better in cases where there were many users and examples that the model was trained with. Having such a dense dataset with many activity labels may not have been the most ideal for this experiment; perhaps using a more generalized dataset might have been better for the goals for this paper. However, the results obtained from the models still show the potential for reliable human activity recognition models using various mobile sensors and machine learning concepts.

The experiments with different models and data from sensors were also limited in more ways than one. Models will be at risk of bias when the training data being fed to the model has uneven amounts of activity labels. Such results can be seen when comparing the results of sitting to running, where the results of running cannot even be considered usable. Using clean labels and the processed data for predictions may also not always be ideal. While it is fast and helped with experimental results and reflecting variable changes, using raw data with some form of feature extraction may be far more ideal for obtaining wanted data points, controlling bias and weights and getting even better experimental results.

VIII. CONCLUSION AND FUTURE WORK

In conclusion, having a variety of sensors is valuable and can improve a model's performance; even if the sensors are limited to those available on consumer devices such as a smartphone or smartwatch. This means that extensive hardware that is often only accessible to the medical industry is not required for human activity recognition systems given the large availability of sensors on consumer devices. The number of sensors in a smartphone is likely enough to produce valuable results using machine learning algorithms; in this case kNN, RF and SVM. A smartwatch may be more limited in results due to having fewer sensors than a smart phone as we have seen from the result differences in accuracy and AUC between groups 1 and the other groups. Having many identical sensors from different mobile devices may not be as beneficial either, as shown from tests that pair identical sensors from different smart devices. This however does show that the sensors available on generally accessible devices are capable enough of achieving acceptable human activity recognition results with potential for improvements in the model. The best results include the model's ability to recognize when a user is sitting with an AUC value of 0.75 when paired with the RF algorithm and Group 4's sensors. Further improvements using more optimized models will be considered and investigated for achieving even more ideal results with the strict limitations of using sensors on a smartphone or smartwatch only.

Other future work includes looking into using raw data from sensors to extract features with more desirable information for training the models. As previously stated, custom feature extraction techniques can be used to feed training data that is far better for models to accurately classify activity labels. Deep learning techniques [17] can be used for both the feature extraction and re-designing the models with a neural network instead of machine learning concepts [18]. There are many features extraction techniques that have been researched and can be potentially utilized in both deep learning and machine learning based models for better results.